\documentclass[twocolumn,10pt]{asme2e}
\usepackage[]{graphicx, epstopdf}
\usepackage{amsmath} 
\usepackage{amsfonts}
\usepackage{amssymb}
\usepackage{bm}
\usepackage{algorithm, algorithmicx}
\usepackage[]{algpseudocode}
\usepackage{color}
\usepackage{subcaption}
\usepackage{caption}
\usepackage{comment}
\usepackage[dvipsnames]{xcolor}
\usepackage[left]{lineno}
\usepackage{float}

\newtheorem{remark}{Remark} 	

\graphicspath{ {figures/} }

\confshortname{IDETC/CIE 2020}
\conffullname{the ASME 2020 International Design Engineering Technical Conferences \&\\
              Computers and Information in Engineering Conference}

\confdate{16-19}
\confmonth{August}
\confyear{2020}
\confcity{St. Louis, MO}
\confcountry{USA}

\papernum{IDETC2020-22722}

\title{Robust Relative Hand Placement for Bi-Manual Tasks}

\author{Anirban Sinha
    \affiliation{
	Department of Mechanical Engineering\\
	Stony Brook University, New York, USA\\
    Email: anirban.sinha@stonybrook.edu
    }	
}

\author{Nilanjan Chakraborty
    \affiliation{
	Department of Mechanical Engineering\\
	Stony Brook University, New York, USA\\
    Email: nilanjan.chakraborty@stonybrook.edu
    }	
}

\begin{document}
\maketitle    
\begin{abstract}
\noindent
{\em In many bi-manual robotic tasks, like peg-in-a-hole assembly, the success of the task execution depends on the error in achieving the desired relative pose between the peg and the hole in a pre-insertion configuration. Random actuation errors in the joint space usually prevents the two arms from reaching their desired task space poses, which in turn results in random error in relative pose between the two hands. This random error varies from trial to trial, and thus depending on the tolerance between the peg and the hole, the outcome of the assembly task may be random (sometimes the task execution succeeds and sometimes it fails). In general, since the relative pose has $6$ degrees-of-freedom, there are an infinite number of joint space solutions for the two arms that correspond to the same task space relative pose. However, in the presence of actuation errors, the joint space solutions are not all identical since they map the joint space error sets differently to the task space. Thus, the goal of this paper is to develop a methodical approach to compute a joint space solution such that the maximum task space error is below a (specified) threshold with high probability. Such a solution is called a  robust inverse kinematics solution for the bi-manual robot. Our proposed method also allows the robot to self-evaluate whether it can perform a given bi-manual task reliably. We use a square peg-in-a-hole assembly scenario on the dual-arm Baxter robot for numerical simulations that shows the utility of our approach.}    
\end{abstract}

\begin{figure}[!htbp]
    \centering
    \includegraphics[width=0.475\textwidth]{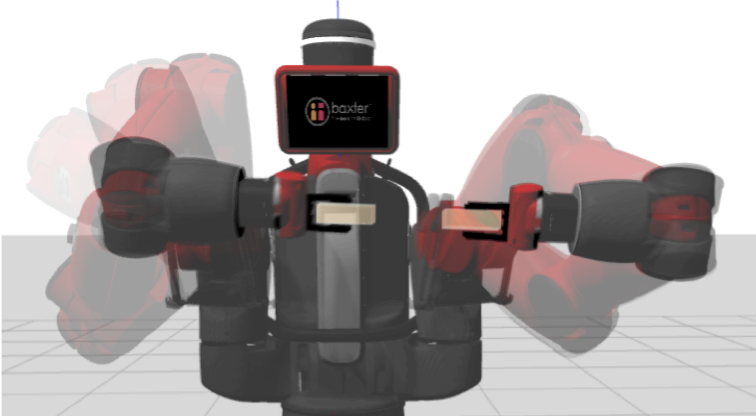}
    \caption{Example pre-insertion configuration in a bi-manual peg-in-a-hole assembly task. There are multiple possible arm joint angles for a given relative pose of the peg and hole.}
    \label{fig: dual_arm_redundancy}
\end{figure}
\section{Introduction}
The use of dual armed or bi-manual manipulation has been envisioned in many applications in environments structured for humans. Example applications include flexible automation and domestic service robotics~\cite{SmithKL+12}. Common bi-manual tasks include object transfer from one hand to other, peg-in-a-hole type assemblies, and transport of an object while holding it with two hands.  Figure~\ref{fig: dual_arm_redundancy} shows a canonical peg-in-a-hole assembly task, which is usually accomplished by (a) moving the hole and the peg to a pre-insertion pose (position and orientation) (b) holding the arm with the hole (or peg) fixed and using an insertion strategy to move the arm with the peg (or hole) towards the hole (or peg) to complete the assembly. 
If both the hole and the peg are moved to the desired pre-insertion pose without any error, then the assembly operation has high chances of being successful. 
However, in practice, joint sensors and actuators come with inherent random errors. This results in imperfect placement of hole and peg. Thus, the actual relative pose between the two hands is different from the desired relative pose. When the available clearance between the peg and the hole is large enough, these errors can still be tolerated for successful assembly. However, for small clearance, achieving successful assembly in the presence of these uncertainties is difficult, and the performance of the robot becomes unreliable. Motivated by the above qualitative discussion, the goal of this paper is to quantitatively study the question of understanding the reliability of performing dual handed tasks. 

For dual-handed manipulation, there are usually infinitely many ways to achieve the desired relative pose between the two hands. This is because the relative pose between two hands is $6$ degree of freedom (DoF) and each of the robot arms usually have at least $6$ DoF. The joint angles for each arm can be obtained by solving the inverse kinematics (IK) problem for each arm. All the IK solutions are equivalent in the absence of joint space error. However, in the presence of joint space error, different IK solutions map joint space error to end-effector (or task space) error differently~\cite{Sinha2019a}.



Therefore we want to solve the {\em robust relative hand placement problem}, which is defined as follows: {\em Given a desired left (right) hand pose, a desired relative pose of right (left) hand with respect to left (right) hand, a joint space error bound, and a tolerance parameter $\epsilon$, compute a joint space configuration of both arms such that the error between the desired relative hand pose and actual relative hand pose is less than $\epsilon$ (with high probability) under any realization of the uncertainties}. The solution to this problem is termed as robust-IK solution, composed of left and right IK pair. We use  robust-IK and robust-IK-pair interchangeably in the rest of the paper. Our problem is a generalization of the problem studied in~\cite{Sinha2019a}, wherein, we presented a method for computing the joint configuration for robustly placing a manipulator arm at a desired pose. Applying the method from~\cite{Sinha2019a} to each arm will give us a placement of the end-effector of each arm with error less than $\epsilon$, but it does not guarantee that the relative pose error of the two hands will be less than $\epsilon$.

Although the robust relative hand placement problem is a feasibility problem, we will formulate and solve the minimization version of the problem, which is more general. The objective is to minimize a task dependent error measure while the constraint is an error ellipsoid obtained by {\em propagating joint space errors of both arms} into task space. The {\em key contribution} of this paper is a novel method for propagating the individual error ellipsoids in the joint space of each arm to a single error ellipsoid in the task space that models the set of (possible) relative poses between the two hands. This is done by formulating the dual arm differential kinematics as a pseudo-single-arm differential kinematics (see Section~\ref{sec: kin_dual_arm}).

This differential kinematics formulation allows us to set up the robust relative hand-placement problem in a manner similar to that of a single arm robust IK as presented in~\cite{Sinha2019a}. Therefore, an optimization formulation similar to that of~\cite{Sinha2019a} can be used. However, now, the optimization variables are the joint angles of both arms. The optimal solution gives joint space configurations for both arms that minimize the maximum error from the desired relative pose (irrespective of the realization of the joint space error). We will call this solution the IK solution with best relative hand placement or simply the {\em best IK solution}. Assuming that the joint space errors are small, the {\em best} IK solution as well as the robust IK solution (if one exists) can be computed by splitting the constrained optimization problem into two independent constrained optimization problems, each of which can then be further simplified to an eigenvalue finding problem~\cite{Sinha2019a}. We also present simulation results with a dual-armed Baxter robot that shows the usefulness of our method in determining feasibility of an assembly scenario of a square peg in a square hole for different error characteristics in the joint space.

\section{Related Work} \label{sec: rel_wrk} \noindent
Related work of our research can be divided into two broader areas, namely, error propagation and analysis in manipulators~\cite{Su1992}, and {\em dual-arm manipulation}~\cite{SmithKL+12}. Positioning error in manipulators are of two basic types, namely {\em static errors} (that can be removed by calibrating the arm) and {\em random errors}. Errors in link-lengths, offset lengths, and/or origin of the joints that are not known precisely, introduce constant biases in end-effector configuration, and are often called \textit{static errors}. They do not change over time and hence can be estimated offline and compensated during calibration process of the robot~\cite{Meggiolaro, Mavroidis1997, Wu1983, Wu1984, Veitschegger1986, Chen1984, Chen1987, mooring1991fundamentals, omodei2001calibration,nof1999handbook}. The second kind of error corresponds to random actuation and sensing errors during task execution. They implicitly affect accuracy in joint rotations which in turn affects accuracy at the end-effector of a manipulator. This second kind of error source in positioning tasks is the motivation behind our work. A group theoretic approach to propagate random joint space actuation error into end-effector space has been presented in~\cite{Wang2006}. In~\cite{Wang2006}, the authors present a method to obtain error covariance at the end of each individual link in closed form due to errors in desired joint configurations. By repeating this procedure sequentially for each link of a manipulator they obtain final error covariance at the end-effector. To capture the effect of large joint errors on error covariance authors in~\cite{Wang2008} presented a second order theory of error propagation.
Our goal here is to find IK solution for a given task corresponding to minimum task space error bound. Thus, unlike~\cite{Wang2006, Wang2008} our method does not rely on individual error samples and corresponding frame by frame error covariance computation. The modeling of joint space and task space error sets in this paper follows from our previous work~\cite{Sinha2019a}. We assume small joint errors along with linearized model of forward position and rotation kinematics to propagate a geometric description of the joint space error set into task space. Obviously~\cite{Wang2006, Wang2008} are more effective if joint space errors are large.

To propagate the joint space error of the individual arms into the relative configuration error between the two hands, we use the relative-Jacobian. A relative-Jacobian relates the relative velocity between the two hands to the joint rates. 
Initial derivation of relative-Jacobian for trajectory generation problem for two cooperating robots can be found in~\cite{lewis1990trajectory, lewis1996trajectory}. Different applications of relative-Jacobian has been mentioned in~\cite{jamisola2015more} and the references therein. Re-derivation of relative-Jacobian in modular-form revealing wrench transformation matrix can be found in~\cite{jamisola2015more} and its application in~\cite{jamisola2015modular}. However none of these papers derive the relative-Jacobian using a product of exponential formulation of the forward kinematics of the manipulators, which is a coordinate-free approach~\cite{MLS94}. We present a novel coordinate-free derivation of the relative-Jacobian by using a pseudo-single arm representation of the dual armed manipulator. This allows for a compact and elegant propagation of the set of possible joint space errors of both arms to the relative configuration space between the two end-effectors.
\section{Mathematical Preliminaries}
\label{sec:math_prelim}
Let $SO(3)$ be the Special Orthogonal group of dimension $3$, which is the space of all rigid body rotations. Let $SE(3)$ be the Special Euclidean group of dimension $3$, which is the space of rigid motions (i.e., rotations and translations). $SO(3)$ and $SE(3)$ are defined as follows~\cite{MLS94}: 
$SO(3) = \{{\bf R} \subset {\mathbb R}^{3 \times 3}| {\bf R}^{\rm T}{\bf R} = {\bf R}{\bf R}^{\rm T} = {\bf I}, |{\bf R}| = 1 \}$,
$ SE(3) =  SO(3) \times {\mathbb R}^3 = \{({\bf p},{\bf R}) | {\bf R} \in SO(3), {\bf p} \in {\mathbb R}^3 \} $
where $|{\bf R}|$ is the determinant of ${\bf R}$ and ${\bf I}$ is a $3 \times 3$ identity matrix.  The set of all end effector or hand poses is called the {\em end effector space} or {\em task space} of the robot and is a subset of $SE(3)$. A task space configuration ${\bf g} \in SE(3)$ can be written either as the pair $({\bf p}, {\bf R})$ or as a $4 \times 4$ homogeneous transformation, i.e., ${\bf g} = \begin{bmatrix}
    {\bf R} & {\bf p} \\
    \bm{0} & 1
\end{bmatrix}$, where $\bm{0}$ is a $1 \times 3$ vector with all components as $0$. 
The twist $\bm{\xi}$ is defined as $\bm{\xi}=\left[-(\bm{\omega}\times{\bf q})^{\rm T} ~\bm{\omega}^{\rm T}\right]^{\rm T}$ for a revolute joint, where $\bm{\omega}$ is the axis of the joint and ${\bf q}$ is any point on that axis. For a prismatic joint, $\bm{\xi}= \left[{\bf v}^{\rm T} ~{\bf 0}^{\rm T}\right]^{\rm T}$, where ${\bf v}$ is a unit vector along the prismatic joint axis. The $6 \times 6$ matrix that transforms twists from one frame to another is represented here as ${\rm Ad}_{\rm g} = \begin{bmatrix}
    {\bf R} & \hat{{\bf p}}{\bf R} \\ \bm{0} & {\bf R}
\end{bmatrix}$
where ${\bf g}$ is the transformation of the frame to which we want to transform the twist\cite[pp.55]{MLS94}. The $\hat{\cdot}$ operator converts a $3\times1$ vector to the corresponding $3\times3$ {\em skew symmetric matrix}.

\noindent
{\bf Unit Quaternion Representation of $SO(3)$}:
Unit quaternions are a singularity free representation of $SO(3)$.
A quaternion is a hyper-complex number, which can be represented by the tuple ${\bf q} = (\eta, \epsilon_x, \epsilon_y, \epsilon_z$) which includes a vector $\bm{\epsilon} \in \mathbb{R}^3$ with components $\epsilon_x$, $\epsilon_y $, $\epsilon_z$ and a scalar $\eta$.
For a unit quaternion $\|{\bf q}\| = 1$. In our paper we extensively make use of vector representation of unit quaternions, ${\bf q} = [\eta \quad \bm{\epsilon}^{\rm T}]^{\rm T}$ with its conjugate ${\bf q}^{*} = {\bf q}^{-1} = [\eta \quad -\bm{\epsilon}^{\rm T}]^{\rm T}$. Rotation about an axis ${\bf \omega}$ with angle $\theta$ is a unit quaternion represented as ${\bf q}({\bf \omega}, \theta) = [\cos \theta/2 \quad {\bf \omega}\sin \theta/2]$. 
\textbf{Quaternion algebra: } We will now describe some rules of quaternion algebra that we have used in our derivations. We define {\em left hand} and {\em right hand compound operator} as ${\bf q}^+ = \begin{bmatrix}
    \eta & -\bm{\epsilon}^{\rm T}  \\
	\bm{\epsilon} & \eta {\bf I} + \hat{\bm{\epsilon}}
	\end{bmatrix}
    \quad
    \textrm{and}
    \quad
    {\bf q}^\oplus = \begin{bmatrix}
    \eta & -\bm{\epsilon}^{\rm T}  \\
	\bm{\epsilon} & \eta{\bf I}-\hat{\bm{\epsilon}}  
	\end{bmatrix}$.
Then the product between quaternions ${\bf q}_1$ and ${\bf q}_2$ is ${\bf q}_1 \otimes {\bf q}_2 = {\bf q}_1^ + {\bf q}_2 = {\bf q}_2 ^\oplus {\bf q}_1$.
We represent a vector ${\bf p}_1$ as a quaternion, $[0 \quad {\bf p}_1^{\rm T}]^{\rm T}$. Rotation of ${\bf p}_1$ about an axis ${\bf \omega}$ and angle $\theta$, can be obtained using quaternion ${\bf q}({\bf \omega}, \theta)$ and its conjugate as
\begin{eqnarray} \nonumber
\label{eq: quat_rot_vector}
	\begin{bmatrix}
		0 \\ {\bf p}_2
	\end{bmatrix}
    = \bf{q}^+(\bf{\omega}, \theta)
    \begin{bmatrix}
		0 \\ {\bf p}_1
	\end{bmatrix}^+
    {\bf q}^{-1}({\bf \omega}, \theta)
    = {\bf q}^+{\bf q}^{-1^\oplus}
    \begin{bmatrix}
		0 \\ {\bf p}_1
	\end{bmatrix}\\ \nonumber
\label{eq: quat_rot_mat}
\text{where} \quad {\bf q}^+{\bf q}^{-1^\oplus} = 
\begin{bmatrix}
	1 & \bm{0}^{\rm T} \\ {\bf 0} & e^{\theta\hat{\bf \omega}}
\end{bmatrix}
=
\begin{bmatrix}
	1 & {\bf 0}^{\rm T} \\ \bf{0} & {\bf R}(\bf{\omega}, \theta)
\end{bmatrix}
\end{eqnarray}
\section{Product of Exponential Formulation for Computing Relative Jacobian}
\label{sec: kin_dual_arm}
\noindent
Let $({\bf g}_{\rm rel})$ be the relative pose of the two end-effector frames and ${\bf J}_{\rm rel}^{\rm s}$ be the relative spatial Jacobian. In this section we will present the product of exponential formulation for computing $({\bf g}_{\rm rel})$ and ${\bf J}_{\rm rel}^{\rm s}$.
The Jacobian matrix, ${\bf J}_{\rm rel}^{\rm s}$ is essential for obtaining the position and orientation error sets in the task space.
To the best of our knowledge, the derivation of the relative Jacobian from the product of exponentials formulation of the relative pose has not appeared elsewhere, and is one of the contributions of the paper.

\begin{figure}
    \centering
    \includegraphics[width=0.5\textwidth]{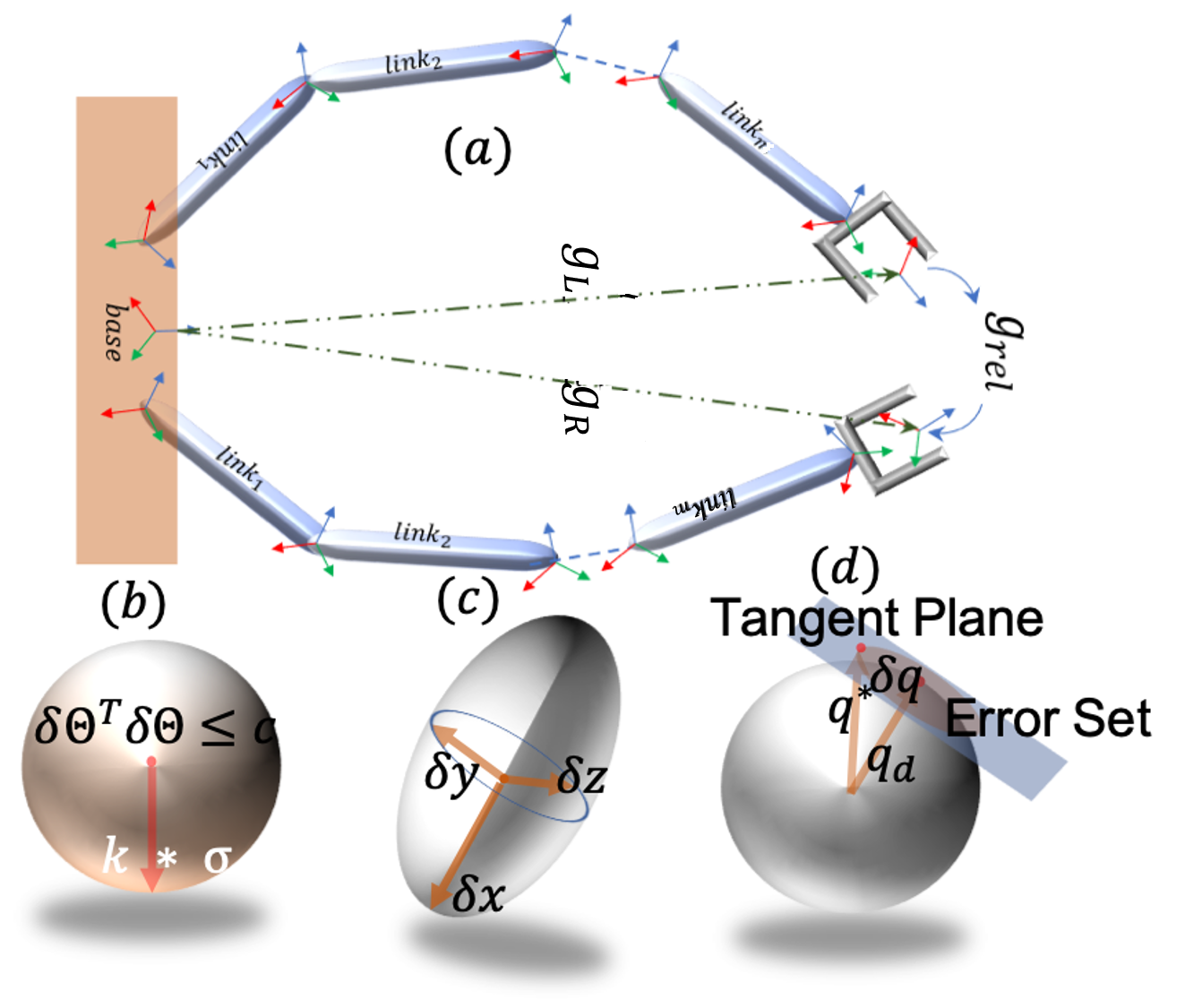}
    \caption{(a) Schematic of a dual arm manipulator. Transform of left and right gripper frames to base frame are ${\bf g}_{\rm dL}$ and ${\bf g}_{\rm dR}$ and ${\bf g}_{\rm rel} = {\bf g}_{\rm dL}^{-1}{\bf g}_{\rm dR}$. (b) joint space error set (shown for $3$ DoF manipulator to visualize). (c) and (d) error-sets to position task space $\in \mathbb{R}^3$ and in tangent space of unit-quaternion sphere $\in SO(3)$.}
    \label{fig: dual_arm_schematic}
\end{figure}

\textbf{Relative hand pose as product of exponentials:}
Let ${\bf g}_{\rm L}, {\bf g}_{\rm R} \in SE(3)$ be the task space poses of the left hand and right hand respectively. The relative task space pose of the right hand with respect to the left hand is ${\bf g}_{\rm rel} = ({\bf p}_{\rm rel}, {\bf R}_{\rm rel})$, where ${\bf p}_{\rm rel}$ is the relative position and ${\bf R}_{\rm rel})$ is the relative orientation.
Suppose, the left arm of the manipulator consists of $\rm n$ joints whereas the right arm has $\rm m$ joints. Let, $\bm{\Theta}_{\rm L} \in \mathbb{R}^{\rm n}$, $\bm{\Theta}_{\rm R} \in \mathbb{R}^{\rm m}$ be the joint solutions such that, ${\bf F}(\bm{\Theta}_{\rm L}) = {\bf g}_{\rm L}$ and ${\bf F}(\bm{\Theta}_{\rm R}) = {\bf g}_{\rm R}$ (see Figure~\ref{fig: dual_arm_schematic}), where ${\bf F}$ is the forward kinematics map. The $i$th joint angle of the left (right) arm is denoted by $\theta_{iL}$ ($\theta_{iR}$) and $i$th joint twist of the left (right) arm is denoted by $\xi_{iL}$ ($\xi_{iR}$).
Using the {\em product of exponential} formula\cite[pp.87]{MLS94} we express ${\bf g}_{\rm L}$ and ${\bf g}_{\rm R}$ as, ${\bf g}_{\rm L}  = \left(\prod_{\rm i=1}^{\rm n} e^{\theta_{\rm iL}\hat{{\bf \xi}}_{\rm iL}}\right) {\bf g}_{\rm L0}$ and ${\bf g}_{\rm R}  = \left(\prod_{\rm i=1}^{\rm m} e^{\theta_{\rm iR}\hat{\bf \xi}_{\rm iR}}\right) {\bf g}_{\rm R0}$.
%
where ${\bf g}_{\rm L0}$ and ${\bf g}_{\rm R0}$ denote left and right hand poses with respect to base frame at the reference configuration (chosen as the one where all joint angles are zero). Then the pose of right hand frame with respect to the left hand frame is (top panel of Figure~\ref{fig: dual_arm_schematic})
\begin{equation} 
\label{eq: g_rel}
\begin{split}
    {\bf g}_{\rm rel}  =  {\bf g}_{\rm L}^{-1}{\bf g}_{\rm R} 
     =  \left(\prod_{\rm i=1}^{\rm n}e^{\hat{\bm{\xi}}_{\rm iL}\theta_{\rm iL}}{\bf g}_{\rm L0}\right)^{-1}\left(\prod_{\rm i=1}^{\rm m}e^{\hat{\bm{\xi}}_{\rm iR}\theta_{\rm iR}}{\bf g}_{\rm R0}\right)\\ 
     =  {\bf g}_{\rm L0}^{-1}\left(\prod_{\rm i=n}^{1}e^{-\hat{\bm{\xi}}_{\rm iL}\theta_{\rm iL}}\right) \left(\prod_{\rm i=1}^{\rm m}e^{\hat{\bm{\xi}}_{\rm iR}\theta_{\rm iR}}\right) {\bf g}_{\rm R0} = {\bf g}_{\rm rel}(\bm{\Theta})
     \end{split}
\end{equation}
The ${\bf g}_{\rm rel}$ in ~\eqref{eq: g_rel} represents the kinematics of a pseudo-single arm with joint sequence ${\rm nL}, \dots, {\rm 1L}, {\rm 1R}, \dots, {\rm mR}$, with joint angle vector $\bm{\Theta}$ obtained by reversing $\bm{\Theta}_{\rm L}$ and concatenating with $\bm{\Theta}_{\rm R}$. Now we derive {\em spatial relative Jacobian matrix} ${\bf J}_{\rm rel}^{\rm s}$ by taking the derivative ${\bf g}_{\rm rel}$ with respect to $\bm{\Theta}$.

\textbf{Computing relative Jacobian:} Since ${\bf g}_{\rm rel}$ is a function of $\bm{\Theta}_{\rm L}$ and $\bm{\Theta}_{\rm R}$, we need to differentiate ${\bf g}_{\rm rel}$ with respect to each element of $\bm{\Theta}_{\rm L}$ and $\bm{\Theta}_{\rm R}$ as in ~\eqref{eq: J_rel_cols}\cite[pp.115]{MLS94} to construct the relative Jacobian matrix.
\begin{equation}
\label{eq: J_rel_cols}
        {\bf J}_{\rm rel}^{\rm s} = 
    \begin{bmatrix}
        \left(\frac{\partial {\bf g}_{\rm rel}}{\partial \theta_{\rm nL}}{\bf g}_{\rm rel}^{-1}\right)^\vee & \cdots & \left(\frac{\partial {\bf g}_{\rm rel}}{\partial \theta_{\rm 1L}}{\bf g}_{\rm rel}^{-1}\right)^\vee & \left(\frac{\partial {\bf g}_{\rm rel}}{\partial \theta_{\rm 1R}}{\bf g}_{\rm rel}^{-1}\right)^\vee & \cdots & \left(\frac{\partial {\bf g}_{\rm rel}}{\partial \theta_{\rm mR}}{\bf g}_{\rm rel}^{-1}\right)^\vee
    \end{bmatrix}
\end{equation}
To obtain the first column of ${\bf J}_{\rm rel}$, i.e., $\left(\frac{\partial {\bf g}_{\rm rel}}{\partial \theta_{\rm nL}}{\bf g}_{\rm rel}^{-1}\right)^\vee$, we differentiate ${\bf g}_{\rm rel}$ in ~\eqref{eq: g_rel} with respect to $\theta_{\rm nL}$ to get
\begin{equation}
    \label{eq: g_rel_partial_1}
    \frac{\partial {\bf g}_{\rm rel}}{\partial \theta_{\rm nL}} = {\bf g}_{\rm L0}^{-1}\left(-\hat{\xi}_{\rm nL}\right) \left(\prod_{\rm i=n}^{1}e^{-\hat{\bm{\xi}}_{\rm iL}\theta_{\rm iL}}\right) \left(\prod_{\rm i=1}^{\rm m}e^{\hat{\bm{\xi}}_{\rm iR}\theta_{\rm iR}}\right) {\bf g}_{\rm R0}
\end{equation}
Right multiplication by ${\bf g}_{\rm rel}^{-1}$ on both sides of ~\eqref{eq: g_rel_partial_1} gives  $\frac{\partial {\bf g}_{\rm rel}}{\partial \theta_{\rm nL}}{\bf g}_{\rm rel}^{-1} = {\bf g}_{\rm L0}^{-1}\left(-\hat{\xi}_{\rm nL}\right){\bf g}_{\rm L0}$.
Then using the $\vee$ operator\cite[pp.115]{MLS94}, twist coordinates are retrieved as,
\begin{equation}
    \label{eq: g_rel_partial_main1}
   \left(\frac{\partial {\bf g}_{\rm rel}}{\partial \theta_{\rm nL}}{\bf g}_{\rm rel}^{-1}\right)^{\vee}  = -{\rm Ad}_{{\bf g}_{\rm L0}^{-1}}\xi_{\rm nL} = -{\rm Ad}^{-1}_{{\bf g}_{\rm L0}}\xi_{\rm nL}
\end{equation}
To derive the expression of $\left(\frac{\partial {\bf g}_{\rm rel}}{\partial \theta_{\rm kL}}{\bf g}_{\rm rel}^{-1}\right)^\vee$, where $\rm k = (m-1), \dots, 1$, we differentiate ${\bf g}_{\rm rel}$ with respect to $\theta_{\rm kL}$ to get
\begin{equation}
    \label{eq: g_rel_partial_3}
    \frac{\partial {\bf g}_{\rm rel}}{\partial \theta_{\rm kL}} = {\bf g}_{\rm L0}^{-1}\left(\prod_{\rm i=n}^{\rm k+1}e^{-\hat{\bm{\xi}}_{\rm iL}\theta_{\rm iL}}\right)\left(-\hat{\bm{\xi}}_{\rm kL}\right)\left(\prod_{\rm i=k}^{1}e^{-\hat{\bm{\xi}}_{\rm iL}\theta_{\rm iL}}\prod_{\rm i=1}^{\rm m}e^{\hat{\bm{\xi}}_{\rm iR}\theta_{\rm iR}}\right) {\bf g}_{\rm R0}
\end{equation}
Multiplying both sides of ~\eqref{eq: g_rel_partial_3} by ${\bf g}_{\rm rel}^{-1}$ and simplifying we get,
\begin{eqnarray} \nonumber
    \label{eq: g_rel_partial_4}
    \frac{\partial {\bf g}_{\rm rel}}{\partial \theta_{\rm kL}}{\bf g}_{\rm rel}^{-1} = {\bf g}_{\rm L0}^{-1}\left(\prod_{\rm i=n}^{\rm k+1}e^{-\hat{\bm{\xi}}_{\rm iL}\theta_{\rm iL}}\right)\left(-\hat{\bm{\xi}}_{\rm kL}\right) \left(\prod_{\rm i=k+1}^{\rm m}e^{\hat{\bm{\xi}}_{\rm iL}\theta_{\rm iL}}\right){\bf g}_{\rm L0}\\
    =\left({\bf g}_{\rm L0}^{-1}\prod_{\rm i=n}^{\rm k+1}e^{-\hat{\bm{\xi}}_{\rm iL}\theta_{\rm iL}}\right)\left(-\hat{\bm{\xi}}_{\rm kL}\right)\left({\bf g}_{\rm L0}^{-1}\prod_{\rm i=n}^{\rm k+1}e^{-\hat{\bm{\xi}}_{\rm iL}\theta_{\rm iL}}\right)^{-1}
\end{eqnarray}
Again using $\vee$ operator, we retrieve the twist coordinate as,
\begin{eqnarray}
    \label{eq: g_rel_partial_main2}
    \left(\frac{\partial {\bf g}_{\rm rel}}{\partial \theta_{\rm kL}}{\bf g}_{\rm rel}^{-1}\right)^{\vee} & = & -{\rm Ad}_{\left({\bf g}_{\rm L0}^{-1}\prod_{\rm i=n}^{\rm k+1}e^{-\hat{\bm{\xi}}_{\rm iL}\theta_{\rm iL}}\right)}\bm{\xi}_{\rm kL}
\end{eqnarray}
To derive the expression of $\left(\frac{\partial {\bf g}_{\rm rel}}{\partial \theta_{\rm kR}}{\bf g}_{\rm rel}^{-1}\right)^\vee$, where ${\rm k = 1, \dots m}$, we differentiate ${\bf g}_{\rm rel}$ with respect to $\theta_{\rm kR}$ to get
\begin{equation}
\label{eq: g_rel_partial_5}
    \frac{\partial {\bf g}_{\rm rel}}{\partial \theta_{\rm kR}} = \left(\prod_{\rm i=1}^{\rm n}e^{\hat{\bm{\xi}}_{\rm iL}\theta_{\rm iL}}{\bf g}_{\rm L0}\right)^{-1}\left(\prod_{\rm i=1}^{\rm k-1}e^{\hat{\bm{\xi}}_{\rm iR}\theta_{\rm iR}}\right)\left(\hat{\bm{\xi}}_{\rm kR}\right)\left(\rm \prod_{\rm i=k}^{\rm m}e^{\hat{\bm{\xi}}_{\rm iR}\theta_{\rm iR}}\right){\bf g}_{\rm R0}
\end{equation}
Multiplying both sides of ~\eqref{eq: g_rel_partial_5} by ${\bf g}_{\rm rel}^{-1}$ we get,
\begin{eqnarray} \nonumber
    \label{eq: g_rel_partial_6}
    \frac{\partial {\bf g}_{\rm rel}}{\partial \theta_{\rm kR}}{\bf g}_{\rm rel}^{-1} = \left(\prod_{\rm i=1}^{\rm n}e^{\hat{\bm{\xi}}_{\rm iL}\theta_{\rm iL}}{\bf g}_{\rm L0}\right)^{-1}\left(\prod_{\rm i=1}^{\rm k-1}e^{\hat{\bm{\xi}}_{\rm iR}\theta_{\rm iR}}\right)\left(\hat{\bm{\xi}}_{\rm kR}\right)\\
    \left(\prod_{\rm i=1}^{\rm k-1}e^{\hat{\bm{\xi}}_{\rm iR}\theta_{\rm iR}}\right)^{-1}\left(\prod_{\rm i=n}^{\rm k+1}e^{\hat{\bm{\xi}}_{\rm iL}\theta_{\rm iL}}{\bf g}_{\rm L0}\right)
\end{eqnarray}
Then, using the $\vee$ operator, we retrieve the twist coordinate as,
\begin{equation}
    \label{eq: g_rel_partial_main3}
    \left(\frac{\partial {\bf g}_{\rm rel}}{\partial \theta_{\rm kR}}{\bf g}_{\rm rel}^{-1}\right)^\vee = {\rm Ad}_{\left(\prod_{i=1}^{n}e^{\hat{\bm{\xi}}_{\rm iL}\theta_{\rm iL}}{\bf g}_{\rm L0}\right)^{-1}\left(\prod_{i=1}^{k-1}e^{\hat{\bm{\xi}}_{\rm iR}\theta_{\rm iR}}\right)}\bm{\xi}_{\rm kR}
\end{equation}
Using ~\eqref{eq: g_rel_partial_main1},~\eqref{eq: g_rel_partial_main2},~\eqref{eq: g_rel_partial_main3} we can find the columns of ${\bf J}_{\rm rel}^{\rm s}$ of ~\eqref{eq: J_rel_cols}. The first and last three rows maps joint space velocities to task space linear and angular velocities respectively. Next we give a brief review of {\em computing robust IK} method presented in~\cite{Sinha2019a} and finally derive the optimization problem of computing robust-IK pair for robust bi-manual hand placement. The summary of computing all the columns of ${\bf J}_{\rm rel}^{\rm s} \in \mathbb{R}^{6 \times \rm (n+m)}$ is given in ~\eqref{eq: j_rel_all_col}.
\begin{eqnarray}
\label{eq: j_rel_all_col}
{\bf J}_{\rm rel}^{\rm s} = 
\left[
\begin{array}{c|c|c}
\left(\frac{\partial {\bf g}_{\rm rel}}{\partial \theta_{\rm nL}}{\bf g}_{\rm rel}^{-1}\right)^\vee & \left(\frac{\partial {\bf g}_{\rm rel}}{\partial \theta_{\rm kL}}{\bf g}_{\rm rel}^{-1}\right)^\vee & \left(\frac{\partial {\bf g}_{\rm rel}}{\partial \theta_{\rm kR}}{\bf g}_{\rm rel}^{-1}\right)^\vee\\
\text{use} & \text{for } \rm k=\rm (n-1) \dots 1 & \text{for } \rm k=1 \dots \rm m\\
 \text{Eq.}\eqref{eq: g_rel_partial_main1} & \text{use Eq.}\eqref{eq: g_rel_partial_main2} & \text{use Eq.}\eqref{eq: g_rel_partial_main3}\\
\end{array}
\right]
\end{eqnarray}
We can then compute {\em analytical Jacobian}, ${\bf J}_{\rm rel}^{\rm a} = \begin{bmatrix}
    {\bf I} & -\hat{\bf p}_{\rm rel} \\ {\bf 0} & {\bf I}
\end{bmatrix}{\bf J}_{\rm rel}^{\rm s}$ where ${\bf p}_{\rm rel}$ is the position vector of ${\bf g}_{\rm rel}$ in~\eqref{eq: g_rel}. Next we derive position and orientation task space error bounds using ${\bf J}_{\rm rel}^{\rm a}$.

\section{Optimization problem for robust bi-manual tasks} \label{sec: robust-IK-pair-formulation}
In bi-manual assembly tasks where robot experiences actuation uncertainties in both the arm joints, success of task depends on the relative placement of the left and right end-effectors  not on the independent placement of the same. Therefore we can not directly use the formulation in~\cite{Sinha2019a} which relied on uncertainty only in one arm. Although the framework to formulate the optimization problem of computing robust-IK-pair for bi-manual tasks is similar to robust-IK problem in~\cite{Sinha2019a}, there are a few modifications needed in representing the error sets and error metrics to be used. Next we first provide the key steps in computing robust-IK method~\cite{Sinha2019a} which we will be adopting and modifying to formulate robust-IK-pair problem for bi-manual tasks with uncertainties in both left and right arms.

Computing robust-IK as described in~\cite{Sinha2019a} has three key steps. (a) Modeling the random joint space error using some probabilistic model. (b) Propagating the joint space error in task space for position and orientation respectively to obtain the error sets in respective spaces. (c) Constructing and solving a min-max constrained optimization problem with an objective chosen using a heuristic error measure. The method computes an inverse kinematics solution by minimizing maximum possible error. We will follow similar steps to formulate the robust-IK-pair problem with two major differences. First, the solution of the robust-IK-pair problem is a concatenated vector of IK-solutions of left and right arm. Second, instead of minimizing the maximum of the absolute position and orientation error as in~\cite{Sinha2019a}, here we minimize the maximum relative position and orientation error. Next we describe the key steps to formulate the robust-IK-pair problem.

\textbf{a. Uncertainty modeling in joint space}: Suppose the left and right right arms are commanded to move to $\bm{\Theta}_{\rm L}\in\mathbb{R}^{\rm n}$ and $\bm{\Theta}_{\rm R}\in\mathbb{R}^{\rm m}$ respectively. Then the nominal commanded concatenated joint vector is $\bar{\bm{\Theta}}\in\mathbb{R}^{\rm n+m}$. The associated joint error due to actuation uncertainties in both left and right arms is denoted as $\bm{\delta \Theta}$. Assuming $\bm{\delta \Theta}$ is normally distributed, i.e., $\bm{\delta \Theta}\sim \mathcal{N}(\bm{0}, \bm{\Sigma})$ where $\bm{\Sigma}=\sigma^2{\bf I}_{\rm (n+m)\times \rm (n+m)}$, the joint space error set is modeled as, $\bm{\delta \Theta}^{\rm T} \bm{\delta\Theta} \leq \rm c$ where, $\rm c = (\rm k \sigma)^2$, 
$\sigma$ is the standard deviation of each joint and $\rm k$ is the number of standard deviation defining radius of the ball shaped joint space error set. We model the joint space error set as a ball by assuming that each joint has the same variance for convenience and ease of presentation. Our method is valid even if the noise variance of each joint is different and/or the joint noises are correlated. 

\textbf{b. Obtaining error sets for position and orientation}: We define the forward position and rotation kinematics map as ${\bf F}_{\rm p}(\bm{\Theta}):\mathbb{R}^{\rm m+n}\rightarrow\mathbb{R}^3$ and ${\bf q}_{\rm r}(\bm{\Theta}):\mathbb{R}^{\rm m+n}\rightarrow SO(3)$. Note that we are using a unit quaternion representation of $SO(3)$. To obtain the error sets in position and orientation task space, the joint space error is propagated to the position and orientation spaces through linearized forward position and orientation kinematics map as shown in~\cite{Sinha2019a}. Here, we will only present the final error equations in~\eqref{eq:error_Taylors_rel} and~\eqref{eq:uncertainty_SO3_rel} respectively. In Figure~\ref{fig: dual_arm_schematic}(b) we have plotted joint space error set (ball) for a $3$DoF robot for visualization. The corresponding propagated the error sets position and orientation task space are visualized in Figure~\ref{fig: dual_arm_schematic}(c) and Figure~\ref{fig: dual_arm_schematic}(d). 
\begin{eqnarray}
\label{eq:error_Taylors_rel}
\bm{\delta}{\bf X}_{\rm rel} = {\bf F}_{\rm p}(\bar{\bm{\Theta}}+ \delta\bm{\Theta}) - {\bf F}_{\rm p}(\bar{\bm{\Theta}}) \approx  {\bf J}_{\rm p_{\rm rel}} \bm{\delta \Theta}\\
\label{eq:uncertainty_SO3_rel}
\bm{\delta}{\bf q}_{\rm rel}  =  \frac{\partial {\bf q}_{\rm r}(\bar{\bm{\Theta}})}{\partial \bm{\Theta}}\bm{\delta}\bm{\Theta} \approx \frac{1}{2}{\bf H}^{\rm T} {\bf J}_{\rm r_{\rm rel}}\bm{\delta\Theta}
\end{eqnarray}
where $\bm{\delta}{\bf X}_{\rm rel}$ is a $3 \times 1$ vector of the relative position error, $\bm{\delta}{\bf q}_{\rm rel}$ is a $4 \times 1$ vector of relative orientation error, and  ${\bf H}({\bf q}) = \left[-\bm{\epsilon} \quad \eta{\bf I} + \hat{\bm{\epsilon}}\right], {\bf I}$ is $3\times3$ identity matrix. The derivation of $\frac{\partial {\bf q}_{\rm r}(\bar{\bm{\Theta}})}{\partial \bm{\Theta}}$ in~\eqref{eq:uncertainty_SO3_rel} from the first principles is given in the Appendix. The error sets for position and rotation task spaces with respect to the relative desired position and orientation are given in ~\eqref{eq:rel_err4},~\eqref{eq:rel_uncertainty_SO3} where ${\bf J}_{\rm p_{\rm rel}}$ and ${\bf J}_{\rm r_{\rm rel}}$ are the first and last three rows of ${\bf J}_{\rm rel}^{\rm a}$, derived in Section~\ref{sec: kin_dual_arm}.
\begin{eqnarray}
\label{eq:rel_err4}
&&\bm{\delta}{\bf X}_{\rm rel}^{\rm T} \left[{\bf J}_{\rm p_{\rm rel}}{\bf J}_{\rm p_{\rm rel}}^{\rm T}\right]^{-1}\bm{\delta}{\bf X}_{\rm rel} \leq c\\
\label{eq:rel_uncertainty_SO3}
&&\bm{\delta}{\bf q}_{\rm rel}^{\rm T} {\bf H}({\bf q}_{\rm rel})^{\rm T} \left[{\bf J}_{\rm r_{\rm rel}} {\bf J}_{\rm r_{\rm rel}}^{\rm T}\right]^{-1}{\bf H}({\bf q}_{\rm rel}) \bm{\delta}{\bf q}_{\rm rel} \leq {\rm c}/4
\end{eqnarray}
\begin{remark}
Since the relative task pose is a subset of $SE(3)$, which is not a vector space, we consider the position and orientation errors separately instead of a combined pose error vector. There is no bi-invariant metric in $SE(3)$ and so defining a notion of distance between the desired and actual pose that applies across all situations is not sensible. As shown below, we use a task-specific weighting to combine the position and orientation error.
\end{remark}

\textbf{c. Robust-IK constrained optimization problem :} Defining the position and orientation error metric as $\mathbb{P}$ and $\mathbb{O}$, the constrained optimization problem yielding robust-IK-pair is,
\begin{eqnarray}\nonumber
	\label{eq: rel_main_opt_problem}
	\underset{\bm{\Theta} \in \mathbb{R}^{\rm n+m}}{\text{argmin\quad}}\underset{\bm{\delta}{\bf X}_{\rm rel}, \bm{\delta}{\bf q}_{\rm rel}}{\text{max\quad}} &  & \mathbb{M} = \mathbb{P} + \gamma \mathbb{O}\\
	\text{subject to           } & & {\bf g}(\bm{\Theta}) = {\bf g}_{\rm rel} \\ \nonumber
	& &\bm{\delta}{\bf X}^{\rm T}_{\rm rel} \left[{\bf J}_{\rm p_{\rm rel}}{\bf J}_{\rm p_{\rm rel}}^{\rm T}\right]^{-1}\bm{\delta}{\bf X}_{\rm rel} \leq {\rm c} \\ \nonumber
	& & {\bf q}^{\rm T}{\bf q} = 1 \\ \nonumber
	& &\bm{\delta}{\bf q}^{\rm T}_{\rm rel}\left[{\bf H}^{\rm T}\left({\bf J}_{\rm r_{\rm rel}}{\bf J}_{\rm r_{\rm rel}}^{\rm T}\right)^{-1}{\bf H}\right]\bm{\delta}{\bf q}_{\rm rel}  \leq  \frac{\rm c}{4}
\end{eqnarray}

\textbf{d. Solution approach :} Here we briefly describe the solution approach outlined in~\cite{Sinha2019a} in the context of the optimization problem in~\eqref{eq: rel_main_opt_problem}. Please note that in ~\eqref{eq: rel_main_opt_problem}, the objective is separable for position and orientation. Also the second constraint is based on ${\bf X}_{\rm rel}$ whereas the third and fourth constraints are based on ${\bf q}_{\rm rel}$ only. Further we can get rid of the first constraint if we restrict our search space only in the IK solution space of both arms. This allows us to split the inner maximization problem of~\eqref{eq: rel_main_opt_problem} into two independent smaller maximization problems as in~\eqref{eq: opti_position_only} and~\eqref{eq: opti_orientation_only}.
\begin{eqnarray}
\label{eq: opti_position_only}
\underset{\bm{\delta}{\bf X}_{\rm rel}}{\text{max\quad}} &  & \mathbb{P}\\ \nonumber
\text{subject to           }
& &\bm{\delta}{\bf X}^{\rm T}_{\rm rel} \left[{\bf J}_{\rm p_{\rm rel}}{\bf J}_{\rm p_{\rm rel}}^{\rm T}\right]^{-1}\bm{\delta} {\bf X}_{\rm rel} \leq {\rm c} \\ 
\label{eq: opti_orientation_only}
\underset{{\bf \delta q}_{\rm rel}}{\text{max\quad}} &  & \mathbb{O}\\ \nonumber
\text{subject to           }
& & {\bf q}^{\rm T} {\bf q} = 1 \\ \nonumber
& &
\bm{\delta}{\bf q}^{\rm T}_{\rm rel} \left[{\bf H}^{\rm T}\left({\bf J}_{\rm r_{\rm rel}}{\bf J}_{\rm r_{\rm rel}}^{\rm T}\right)^{-1}{\bf H}\right]\bm{\delta}{\bf q}_{\rm rel}  \leq  \frac{\rm c}{4}
\end{eqnarray}
For a given IK solution of both arms as a concatenated vector $\bm{\Theta}$, the maximized objective values $\mathbb{P}^*$ and $\mathbb{O}^*$ can be obtained by solving for maximum eigenvalues of the characteristic matrices of the position and orientation error sets. Detailed reasoning on why the smaller optimization problems in ~\eqref{eq: opti_position_only} and~\eqref{eq: opti_orientation_only} can be posed as eigenvalue finding problem can be found in~\cite{Sinha2019a}. We just present the final expressions of maximum $\mathbb{P}=\mathbb{P}^*$ and maximum $\mathbb{O}=\mathbb{O}^*$ for a given value of $\bm{\Theta}$.
\begin{eqnarray}
    \label{eq: main_position_opti}
    \mathbb{P}^* & = & \underset{\lambda}{\text{max\quad}} \text{eig}\left({\rm c}\left[{\bf J}_{\rm p_{\rm rel}}(\bm{\Theta}){\bf J}_{\rm p_{\rm rel}}(\bm{\Theta})^{\rm T}\right]\right)\\
    \label{eq: rotation_error_measure}
    \mathbb{O}^* & = & \arccos {\bf q}_{\rm rel}^{\rm T}{\bf q}^*
\end{eqnarray}
where ${\bf q}^* = ({\bf q}_{\rm rel} + {\bf H}^{\rm T} {\bf v}^*)/||{\bf q}_{\rm rel} + {\bf H}^{\rm T} {\bf v}^*||$ and ${\bf v}^* = \frac{1}{2}\sqrt{\rm c \lambda_{\rm max}}{\bf V}_{\rm max}$, where $\lambda_{\rm max}$ is maximum eigenvalue of ${\bf J}_{\rm r_{\rm rel}}{\bf J}_{\rm r_{\rm rel}}^{\rm T}$ and ${\bf V}_{\rm max}$ is the eigenvector associated to $\lambda_{\rm max}$. Knowing $\mathbb{P}^*$ and $\mathbb{O}^*$, we can compute the weighted metric as $\mathbb{M}^* = \mathbb{P}^* + \gamma \mathbb{O}^*$. We need to repeatedly compute $\mathbb{M}^*$ in this manner for different IK-pair. The IK-pair for which $\mathbb{M}^*$ is minimized, that IK-pair is called the best-IK-pair and is denoted as $\bm{\Theta}^*$. In the above discussion, we are assuming that there is a method available for computing all the IK solutions for each arm. For example, for the Baxter arm we can use the existing IK-solver~\cite{sinha2019geometric}. We are providing a method to select the IK solution pair (if one exists) that can achieve the peg-in-hole assembly robustly. 
\section{Numerical Examples}
\label{sec: num_ex}
\begin{figure}[!htb]
    \centering
    \includegraphics[width=0.47\textwidth]{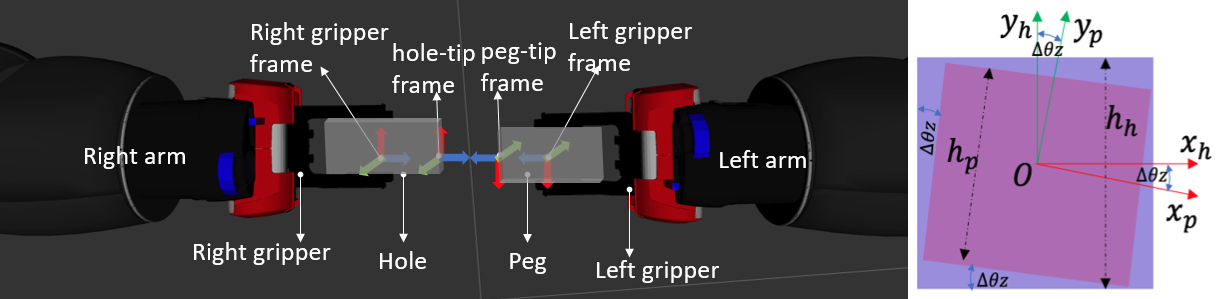}
    \caption{Left: Simulation setup: Left and right end-effectors holding square peg and square hole. The relevant frames ($\rm X, Y, Z$ axes as red, green, blue arrows) to measure relative pose error between peg and hole. Right: Effect of orientation error manifested as error in matching corners of peg (red square) and hole (blue square). This error is captured by the last term in error measure expression, {\em i.e.} $\rm h_{\rm p}\sin{\Delta\theta_{\rm z}}$, discussed in section~\ref{sec: num_ex}.}
    \label{fig: frames}
    \centering
    \includegraphics[scale=0.2, width=9cm, height=3.3cm]{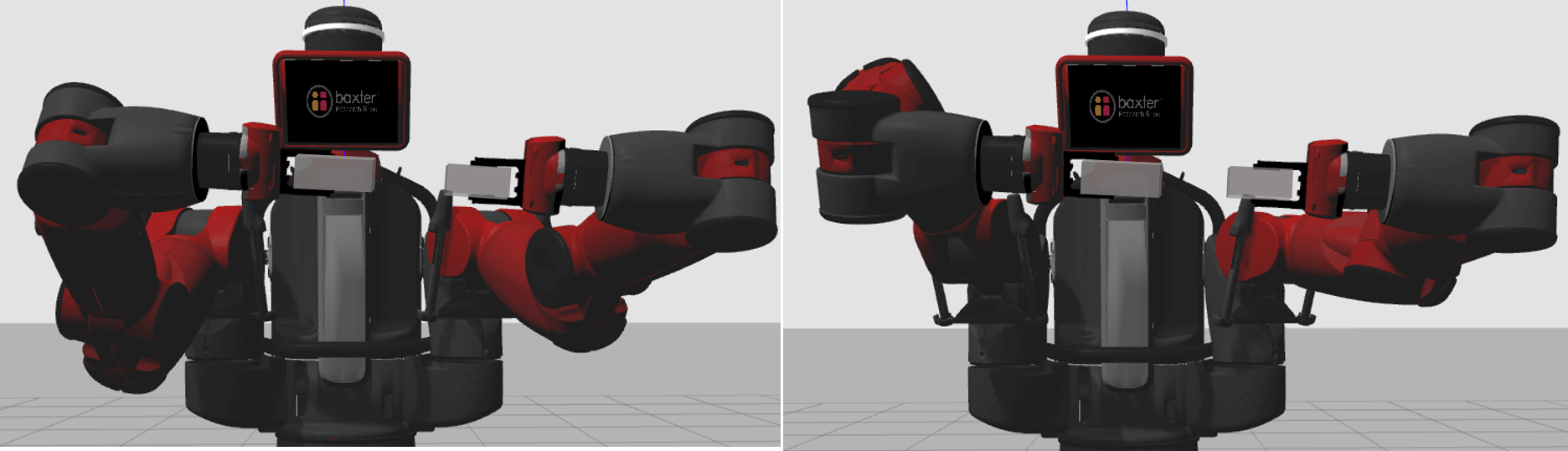}
    \caption{Instances of successful (left panel) and failed (right panel) situations while performing peg in to hole assembly task by executing IK-pairs $\bm{\Theta}^*$ (left panel) and $\bm{\Theta}^-$(right panel).}
    \label{fig: best_worst_simulation}
\end{figure}
\begin{figure}[!htb]
    \centering
    \includegraphics[scale=0.6, width=8cm, height=8cm]{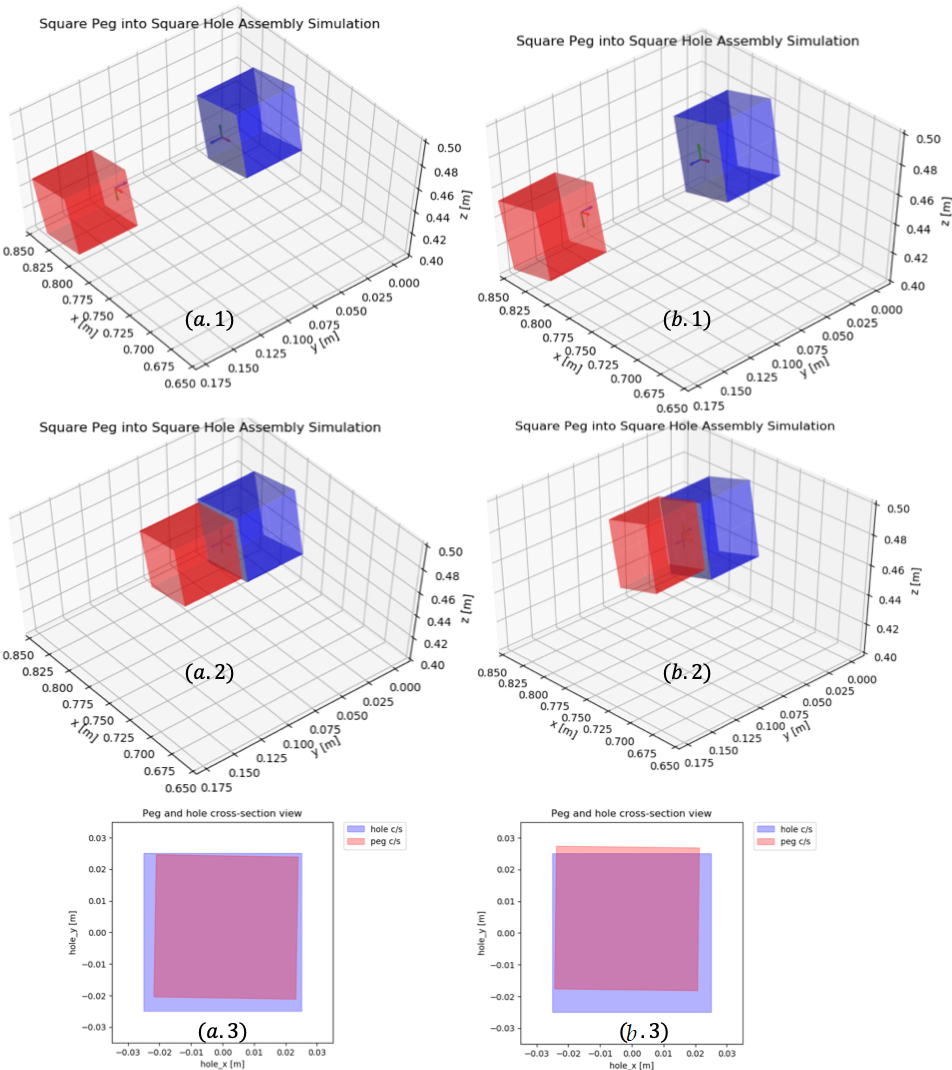}
    \caption{Success $(a)$ and failure $(b)$ instances: Initial $(1)$ and final $(2)$ poses of peg(red) and hole(blue). $(3)$ Projection of peg and hole cross sections on to $\rm XY$ plane of {\em hole\_frame}.}
    \label{fig: success_failure_visualization}
\end{figure}
In this example we consider an application where the robot has to perform a peg-in-a-hole assembly with square cross section. We show that computing robust-IK-pair is useful in placement of left and right grippers with minimal error to a pre-insertion pose. We also perform comparison of success rates using robust-IK-pair and some other IK-pair to show that robust-IK-pair performs the task more reliably. We chose square peg and hole over the circular one because for the prior, both position and orientation error affect the  success of the assembly. All the results presented in this section is obtained using a simulated Baxter robot. We also present results and discussion on how computing robust-IK-pair may endow a robot the capability to understand the feasibility of a given task.

The simulation setup is shown in Figure~\ref{fig: frames}, where the robot holds a square peg and a hole objects in left and right end-effectors. Let $\bar{{\bf g}}_{\rm rel}$ be the transform of {\em hole\_tip\_frame} to {\em peg\_tip\_frame} and ${\bf g}_{\rm rel}$ is the relative transform between left and right end-effectors. If $^{\rm l}{\bf g}_{\rm p}$ and $^{\rm r}{\bf g}_{\rm h}$ are the transforms of peg and hole tip frames with respect to {\em left} and {\em right} gripper frames, then we can write $\bar{{\bf g}}_{\rm rel}$ in terms of ${\bf g}_{\rm rel}$ as, $\bar{{\bf g}}_{\rm rel} = {^{\rm l}{\bf g}_{\rm p}^{-1}} {\bf g}_{\rm rel} {^{\rm r}{\bf g}_{\rm h}}$. The expressions of $^{\rm l}{\bf g}_{\rm p}$ and $^{\rm r}{\bf g}_{\rm h}$ are: $^{\rm l}{\bf g}_{\rm p} =
    \begin{bmatrix}
        {\bf I} & {\bf p}_{\rm p}\\
        \bm{0} & 1
    \end{bmatrix}$, ${^{\rm r}{\bf g}}_{\rm h} =
    \begin{bmatrix}
        {\bf I} & {\bf p}_{\rm h}\\
        \bm{0} & 1
    \end{bmatrix}$, ${\bf p}_{\rm i} = 
    \begin{bmatrix}
        0 & 0 & {\rm l}_{\rm i}
    \end{bmatrix}^{\rm T}$
where ${\rm i} \in \{\rm{p,h}\}$ and ${\rm l}_{\rm p}$, ${\rm l}_{\rm h}$ are the lengths of peg and hole tips from the respective gripper frames. Also let, ${\bf p}_{\rm rel}$, $\bar{{\bf p}}_{\rm rel}$ be the position vectors and ${\bf R}_{\rm rel}$, $\bar{{\bf R}}_{\rm rel}$ be the rotation matrices of ${\bf g}_{\rm rel}$ and $\bar{{\bf g}}_{\rm rel}$ respectively. Further, let ${\bf R}_{\rm rel}^{\rm z}$ be the third column of ${\bf R}_{\rm rel}$. Therefore $\bar{{\bf p}}_{\rm rel}$ is $\bar{{\bf p}}_{\rm rel} = {\bf p}_{\rm rel} + {\rm l}_{\rm h}{\bf R}_{\rm rel}^{\rm z} - {\bf p}_{\rm h}$.
Using superscripts $\rm a$ and $\rm d$ to indicate desired and achieved values, we define error measure, $||\bar{{\bf p}}_{\rm rel}^{\rm d} - \bar{{\bf p}}_{\rm rel}^{\rm a}|| = ||{\bf p}_{\rm rel}^{\rm d} - {\bf p}_{\rm rel}^{\rm a} + {\rm l}_{\rm h}({^{\rm d}{\bf R}_{\rm rel}^{\rm z}} - {^{\rm a}{\bf R}_{\rm rel}^{\rm z}}) + {\rm h}_{\rm p}\sin{\Delta\theta_{\rm z}}||$.
The term ${\bf p}_{\rm rel}^{\rm d} - {\bf p}_{\rm rel}^{\rm a}$ indicates position error of the {\em right\_gripper} with respect to {\em left\_gripper} and can be obtained by solving the sub-problem in ~\eqref{eq: main_position_opti}. The term ${\rm l}_{\rm h}({^{\rm d}{\bf R}_{\rm rel}^{\rm z}} - {^{\rm a}{\bf R}_{\rm rel}^{\rm z}})$ transforms orientation error in $\rm z$-axis of ${\bf R}_{\rm rel}$ into position error and can be computed by solving the sub-problem in ~\eqref{eq: rotation_error_measure}. The last term of error measure expression, {\em i.e.} ${\rm h}_{\rm p}\sin{\Delta\theta_{\rm z}}$, indicates position error of the peg's corner points from its matching hole's corner points where ${\rm h}_{\rm p}$ is the height of the peg and $\Delta\theta_{\rm z}$ is the angular difference between $^{\rm d}{\bf R}_{\rm rel}^{\rm x}$ and $^{\rm a}{\bf R}_{\rm rel}^{\rm x}$ (see right panel of Figure~\ref{fig: frames}). We minimize the error measure to get the robust-IK-pair by iterating over all the combinations of the IK-pairs of left and right arm. While generating the results we assumed that desired peg and hole frame configurations with respect to base frame are:
\begin{eqnarray}
    \label{eq: g_bp}
     {^{\rm b}{\bf g}_{\rm p}} & = &
    \begin{bmatrix}
        -0.976 & -0.212 & -0.045 & 0.752\\
         0.054 & -0.036 & -0.997 & 0.173\\
         0.210 & -0.977 & 0.047 &  0.451\\
        0 & 0 & 0 & 1
    \end{bmatrix}\\
    \label{eq: g_bh}
    {^{\rm b}{\bf g}_{\rm h}} & = &
    \begin{bmatrix}
        -0.976 & 0.212 & 0.045 & 0.748\\
         0.054 & 0.036 & 0.997 & 0.072\\
         0.210 & 0.977 & -0.047 & 0.457\\
        0 & 0 & 0 & 1
    \end{bmatrix}
\end{eqnarray}
\begin{figure}[!htb]
    \centering
    \includegraphics[scale=0.6]{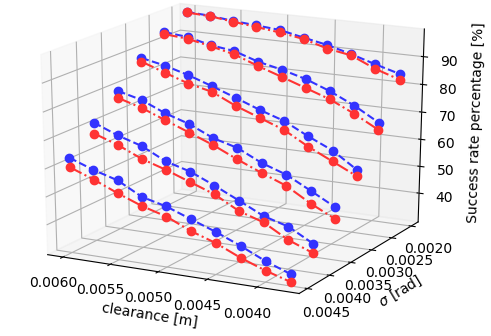}
    \caption{Performance of $\bm{\Theta}^*$ (blue) and $\bm{\Theta}^-$ (red) solutions with varying joint uncertainty $(\sigma)$.} 
    \label{fig: success_varying_std}
    \centering
    \includegraphics[scale=0.5]{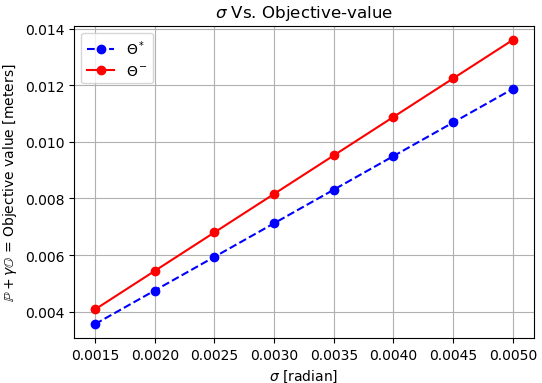}
    \caption{Change in objective value with varying standard deviation $\sigma$ of joint errors for IK-pair $\bm{\Theta}^*$ and $\bm{\Theta}^-$ respectively.}
    \label{fig: feasibility}
\end{figure}
The position vectors of ${^{\rm b}{\bf g}_{\rm p}}$ and ${^{\rm b}{\bf g}_{\rm h}}$ are in meters. Further, we consider both the peg and hole lengths, {\em i.e.}, ${\rm l}_{\rm p}, {\rm l}_{\rm h}$ as $\rm 0.050$ m. Using~\eqref{eq: g_bp} and~\eqref{eq: g_bh} along with known transforms $^{\rm l}{\bf g}_{\rm p}$, we can compute the desired transforms of left and right gripper frames ${\bf g}_{\rm L}$ and ${\bf g}_{\rm R}$. Knowing ${\bf g}_{\rm L}$ and ${\bf g}_{\rm R}$, we can further compute ${\bf g}_{\rm rel} = {\bf g}_{\rm L}^{-1}{\bf g}_{\rm R}$. Then using the proposed method, we can find the best or robust-IK-pair that minimizes the error measure.\\
The robust-IK-pair is found to be $\bm{\Theta}^*\equiv(\bm{\Theta}^{*}_{\rm L}, \bm{\Theta}^{*}_{\rm R})$ where $\bm{\Theta}^{*}_{\rm L}$ $=$ $[-0.362,$ $0.321,$ $-2.994,$ $0.572,$ $1.279,$ $1.932,$ $-0.494]$radian, $\bm{\Theta}^{*}_{\rm R}$ $=$ $[0.494,$ $0.551,$ $2.881,$ $1.210,$ $-1.367,$ $1.552,$ $0.840]$radian and associated objective value is $0.0079$m when $\sigma=0.0045$ and $\rm k=2$ is considered to compute $\rm c$ value while modeling joint space uncertainty.

To compare the performance of $\bm{\Theta}^*$ with some other IK-pair, we consider $\bm{\Theta}^-\equiv(\bm{\Theta}^{-}_{\rm L}, \bm{\Theta}^{-}_{\rm R})$ where $\bm{\Theta}^{-}_{\rm L}$ $=$ $[-0.120,$ $ 0.084,$ $-1.980,$ $0.507,$ $0.324,$ $1.810,$ $-0.347]$radian, $\bm{\Theta}^{-}_{\rm R}$ $=$ $[0.278,$ $-0.710,$ $0.710,$ $1.203,$ $-2.090,$ $-1.336,$ $3.050]$radian. The objective value corresponding to $\bm{\Theta}^-$ is found to be $0.0093$m (see Figure~\ref{fig: best_worst_simulation}). Statistically the objective values indicate that after executing the IK-pairs, the resulting relative error at the {\em hole\_frame} with respect to {\em peg\_frame} will be less than that of the objective value most of the times.

To compare the performance of the IK-pairs $\bm{\Theta}^*$ and $\bm{\Theta}^{-}$, we followed the strategy as illustrated in Figure~\ref{fig: success_failure_visualization}. In the presence of joint uncertainty, we first execute IK-pair $\bm{\Theta}^{*}$ $(\text{or } \bm{\Theta}^{-})$ and get peg and hole poses as in Figure~\ref{fig: success_failure_visualization}$(a.1)$ (or $(b.1)$). Next the peg is moved towards hole along the $\rm z$-axis of the {\em peg\_frame} as shown in Figure~\ref{fig: success_failure_visualization}$(a.2)$ (or $(b.2)$). Then we project square face of the peg in the $\rm XY$ plane of {\em hole\_frame} as in Figure~\ref{fig: success_failure_visualization}$(a.3)$ (or $(b.3)$). If all the vertices of peg lie inside hole's square cross-sectional area, then we count positioning of the peg and hole as successful or failed otherwise. We performed these steps repeatedly for $10000$ times with random joint noise for a particular IK-pair for a given value of clearance and $\sigma$. In Figure~\ref{fig: success_varying_std} we have plotted percentage of success rates for the IK-pair $\bm{\Theta}^{*}$ and $(\bm{\Theta}^{-})$ respectively for varied  joint space uncertainties and clearance between peg and hole ($\sigma$ varying from $0.0020 - 0.0045$ radian and clearance varying from $0.004-0.006$ m). If widths of peg and hole are denoted as ${\rm W}_{\rm p}$ and ${\rm W}_{\rm h}$ respectively, then available clearance is computed as $({\rm W}_{\rm h} - {\rm W}_{\rm p})/2$. 

It can be noticed that the robust-IK-pair $(\bm{\Theta}^{*})$ performed better than $\bm{\Theta}^-$ in terms of success rates for all the combinations of clearance and $\sigma$ considered. Figure~\ref{fig: feasibility} is useful from the perspective of {\em feasibility} analysis of accomplishing peg-in-hole task. For Figure~\ref{fig: feasibility}, the clearance between peg and hole is considered as $0.006$ m and joint space uncertainty is chosen as $\sigma=0.0025$ radian. Notice in Figure~\ref{fig: feasibility}, both $\bm{\Theta}^*$ and $\bm{\Theta}^-$ can perform assembly with high confidence since the objective value lower than $0.006$ m for both the IK-pair. In this case no optimization is required at all. This is also evident from Figure~\ref{fig: success_varying_std} where the success rates of $\bm{\Theta}^*$ at $clearance = 0.006$ m and $\sigma=0.0020$ radian, is almost same. However, if the clearance is $0.008$ m and $\sigma=0.0030$ radian, then $\bm{\Theta}^*$ can accomplish the assembly with higher confidence than $\bm{\Theta}^-$, since objective value for $\bm{\Theta}^*$ is much lower than $0.008$ m. In this scenario, using the proposed method to characterize IK-pair will be useful to accomplishing a task more reliably.
\section{Conclusion}
\label{sec: conclusion}
We have presented a method for computing IK-pair for redundant dual-arm robots to perform peg-in-a-hole type of assembly tasks in a more reliable manner in the presence of joint positioning uncertainty. Using the fact that, success of peg-in-a-hole type tasks is dependent on relative hand placement of dual arm robots, we formulated the best-IK-pair (or robust-IK-pair) problem in terms relative configuration errors. This approach allowed us to represent the kinematics of a dual-arm manipulator as a pseudo-single arm kinematics. Since the Jacobian plays a key role in mapping the joint space error sets to the task space, we derived the Jacobian matrix of the pseudo-single kinematic chain arm using a product of exponential formulation for computing the relative pose. We used the robust-IK constrained optimization problem in~\cite{Sinha2019a} to compute robust-IK-pair for bi-manual tasks. Using a simulated Baxter robot we showed  that using robust-IK-pair, the task of inserting a square peg into a square hole was accomplished more reliably than executing a non-optimal IK-pair. We further showed that the method of computing best or robust-IK-pair can be used to endow robots with capability to self determine feasibility of a given task for a given allowable error margin.  
In future work, we plan to perform experiments with real robot to achieve peg-in-hole task using dual-arm using robust-IK-pair to see the degree of robustness that can achieved in practice. We also plan to model a more generic joint space error model that can capture non-Gaussian error distribution and derivation of corresponding task space error sets. 
\bibliographystyle{asmems4}
\bibliography{dual_arm_ref}
\section*{Appendix} \noindent
This section derives the term $\frac{\partial {\bf q}_{\rm r}(\bar{\bm{\Theta}})}{\partial \bm{\Theta}}$ from ~\eqref{eq:uncertainty_SO3_rel}.
We present this here for completeness, since we could not find any previous work on deriving the partial derivative of the unit quaternion representing relative orientation of the two hands of a dual-armed manipulator with respect to the joint angles of the two manipulators.
Let the orientation of left and right end-effectors with no joint rotations be ${\bf q}_{\rm 0L}$, ${\bf q}_{\rm 0R}$. Then resultant orientation of left and right end-effector are ${\bf q}_{\rm L}$ and ${\bf q}_{\rm R}$ respectively which can be obtained by compounding elementary rotations of each joint as,
\begin{eqnarray} \nonumber
    &&{\bf q}_{\rm L}(\bm{\Theta}_{\rm L}) = {\bf q}_{\rm 1L}(\bm{\omega}_{\rm 1L}, \theta_{\rm 1L})\otimes \cdots \otimes {\bf q}_{\rm nL}(\bm{\omega}_{\rm nL}, \theta_{\rm nL}) \otimes {\bf q}_{\rm 0L}\\
    &&={\bf q}_{\rm 1L}^+ {\bf q}_{\rm 2L}^+ \cdots \bm{q}_{\rm (n-1)L}^+ {\bf q}_{\rm nL}^+ {\bf q}_{\rm 0L}\\ \nonumber
    &&{\bf q}_{\rm R}(\bm{\Theta}_{\rm R}) ={\bf q}_{\rm 1R}(\bm{\omega}_{\rm 1R}, \theta_{\rm 1R})\otimes \cdots \otimes {\bf q}_{\rm mR}(\bm{\omega}_{\rm mR}, \theta_{\rm mR}) \otimes {\bf q}_{\rm 0R}\\
    && = {\bf q}_{\rm 1R}^+ {\bf q}_{\rm 2R}^+ \cdots {\bf q}_{\rm (m-1)R}^+ {\bf q}_{\rm mR}^+{\bf q}_{\rm 0R}
\end{eqnarray}
The relative orientation between left and right end-effector is,
\begin{equation}
    \label{eq: q_rel_derivation}
    {\bf q}_{\rm rel} (\bm{\Theta}) = {\bf q}_{\rm L}^{-1} \otimes {\bf q}_{\rm R} = {\bf q}_{\rm 0L}^{-1} \otimes \left(\prod_{i=n}^1 {\bf q}_{\rm iL}^{-1}\right) \otimes \left(\prod_{j=1}^m {\bf q}_{\rm jr}\right) \otimes {\bf q}_{\rm 0R}
\end{equation}
In matrix multiplication form we can write ${\bf q}_{\rm rel}$ as ${\bf q}_{\rm rel}(\bm{\Theta}) = {\bf q}_{\rm 0L}^{-1^{+}}\left(\prod_{\rm i=n}^1 {\bf q}_{\rm iL}^{-1^+}\right)\left( \prod_{\rm j=1}^{\rm m} {\bf q}_{\rm jr}^+\right){\bf q}_{\rm 0R}$
Notice that ${\bf q}(\bm{\omega}, \theta) = \left[\cos\frac{\theta}{2}, \bm{\omega}\sin\frac{\theta}{2}\right]^{\rm T}$ then ${\bf q}^{-1}(\bm{\omega}, \theta) = \left[\cos\frac{\theta}{2}, -\bm{\omega}\sin\frac{\theta}{2}\right]^{\rm T}$. \cite{Barfoot2011} showed that derivative of such quaternion with respect to the rotation angle $\theta$ is, $\frac{\partial {\bf q}({\bf \omega}, \theta)}{\partial \theta} = \frac{1}{2}
    \begin{bmatrix}
    0 \\ {\bf \omega}
    \end{bmatrix}^+
    {\bf q}(\bf \omega, \theta)$,
Following that derivative of ${\bf q}^{-1}$ with $\theta$ can be obtained as, 
\begin{equation}
\label{eq: inv_quat_partial}
    \frac{\partial {\bf q}^{-1}(\bm{\omega}, \theta)}{\partial \theta} = -\frac{1}{2}
    \begin{bmatrix}
    0 \\ \bm{\omega}
    \end{bmatrix}^+
    {\bf q}^{-1}(\bm{\omega}, \theta)
\end{equation}
The error-quaternion $\bm{\delta}{\bf q}_{\rm rel}$ can be written as $\bm{\delta}{\bf q}_{\rm rel} = \frac{\partial {\bf q}_{\rm rel}}{\partial \bm{\Theta}}\bm{\delta}\bm{\Theta}$, where $\frac{\partial {\bf q}_{\rm rel}}{\partial \bm{\Theta}} =
    \begin{bmatrix}
        \frac{\partial {\bf q}_{\rm rel}}{\partial \bm{\theta}_{\rm nL}} & \cdots & \frac{\partial {\bf q}_{\rm rel}}{\partial {\bf \theta}_{\rm 1L}} & \frac{\partial {\bf q}_{\rm rel}}{\partial {\bf \theta}_{\rm 1R}} & \cdots & \frac{\partial {\bf q}_{\rm rel}}{\partial {\bf \theta}_{\rm mR}}
    \end{bmatrix}$.
Next we derive expressions of partial derivative terms of the columns of $\frac{\partial {\bf q}_{\rm rel}}{\partial\bm{\Theta}}$. We do it by considering three different cases, which can be used to fill up all of the $\rm n+m$ columns of $\frac{\partial {\bf q}_{\rm rel}}{\partial \bm{\Theta}}$. At first we derive the expression of $\frac{\partial \bf q_{\rm rel}}{\partial \theta_{\rm nL}}$ as follows,
\begin{equation}
\label{eq: partial_nl1}
    \frac{\partial {\bf q}_{\rm rel}}{\partial \theta_{\rm nL}} = 
    {\bf q}_{\rm 0L}^{-1} \otimes
    \begin{bmatrix}
        \frac{\partial {\bf q}_{\rm nL}^{-1}}{\partial \theta_{\rm nL}}
    \end{bmatrix}
    \otimes \left(\prod_{\rm i=n-1}^1 {\bf q}_{\rm iL}^{-1}\right)
    \otimes \left(\prod_{\rm j=1}^m {\bf q}_{\rm jR}\right)
    \otimes {\bf q}_{\rm 0R}
\end{equation}
Using the identity in ~\eqref{eq: inv_quat_partial} in to ~\eqref{eq: partial_nl1} we get,
\begin{eqnarray} \nonumber
    &&\frac{\partial {\bf q}_{\rm rel}}{\partial \theta_{\rm nL}} =
    {\bf q}_{\rm 0L}^{-1} \otimes -\frac{1}{2}
    \begin{bmatrix}
        0 \\ \bm{\omega}_{\rm nL}
    \end{bmatrix}
    \otimes \left(\prod_{\rm i=n}^1 {\bf q}_{\rm iL}^{-1}\right)
    \otimes \left(\prod_{\rm j=1}^m {\bf q}_{\rm jR}\right)
    \otimes {\bf q}_{\rm 0R}\\ \nonumber
    && =
    {\bf q}_{\rm 0L}^{-1} \otimes -\frac{1}{2}
    \begin{bmatrix}
        0 \\ \bm{\omega}_{\rm nL}
    \end{bmatrix} {\bf q}_{\rm 0L} \otimes {\bf q}_{\rm 0L}^{-1} \otimes \left(\prod_{\rm i=n}^1 {\bf q}_{\rm iL}^{-1}\right)
    \otimes \left(\prod_{\rm j=1}^m {\bf q}_{\rm jR}\right)
    \otimes {\bf q}_{\rm 0R}\\ \nonumber
    & = &
    -\frac{1}{2} {\bf q}_{\rm 0L}^{-1} \otimes
    \begin{bmatrix}
        0 \\ \bm{\omega}_{\rm nL}
    \end{bmatrix}
    {\bf q}_{\rm 0L}
    \otimes
    {\bf q}_{\rm rel}
    =
    -\frac{1}{2} {\bf q}_{\rm 0L}^{-1^+}{\bf q}_{\rm 0L}
    \begin{bmatrix}
        0 \\ \bm{\omega}_{\rm nL}
    \end{bmatrix}
    {\bf q}_{\rm rel}\\ 
    & = & -\frac{1}{2}
    \begin{bmatrix}
        1 & \bm{0}^{\rm T} \\ 0 & {\bf R}_{\rm 0L}^{-1}
    \end{bmatrix}
    \begin{bmatrix}
        0 \\ \bm{\omega}_{\rm nL}
    \end{bmatrix}
    \otimes {\bf q}_{\rm rel}
    = -\frac{1}{2}
    \begin{bmatrix}
        0 \\ {\bf R}^{\rm T}_{\rm 0L}\bm{\omega}_{\rm nL}
    \end{bmatrix}\otimes {\bf q}_{\rm rel}\\ \nonumber
    & = & -\frac{1}{2} {\bf q}_{\rm rel}^\oplus
    \begin{bmatrix}
        0 \\ {\bf R}^{\rm T}_{\rm 0L}\bm{\omega}_{\rm nL}
    \end{bmatrix} \text{using the right-hand compound operator}
    \label{eq: partial_nl2}
\end{eqnarray}
\noindent
Now we derive a general expression of $\frac{\partial {\bf q}_{\rm rel}}{\partial \theta_{\rm kL}}$ for $\rm k$ = \rm (n-1) \dots $\rm 1$.
\begin{equation}
\begin{split}
    \frac{\partial {\bf q}_{\rm rel}}{\partial \theta_{\rm kL}} =
    \left({\bf q}_{\rm 0L}^{-1}\otimes\prod_{\rm i=n}^{\rm k+1} {\bf q}_{\rm iL}^{-1}\right)\otimes-\frac{1}{2}
    \begin{bmatrix}
        0 \\ \bm{\omega}_{\rm kL}
    \end{bmatrix}\\
    \otimes
    \left(\prod_{\rm i=k}^{1} {\bf q}_{\rm iL}^{-1}\right)
    \otimes
    \left(\prod_{\rm j=1}^{\rm m} {\bf q}_{\rm jR}\right)
    \otimes {\bf q}_{\rm 0R}
\end{split}
\end{equation}
Introducing identity quaternion in the above equation we get,
\begin{equation}
\label{eq: partial_kl}
\begin{split}
    \frac{\partial {\bf q}_{\rm rel}}{\partial \theta_{\rm kL}} =
    \left({\bf q}_{\rm 0L}^{-1}\otimes\prod_{\rm i=n}^{\rm k+1} {\bf q}_{\rm iL}^{-1}\right)\otimes-\frac{1}{2}
    \begin{bmatrix}
        0 \\ {\bf \omega}_{\rm kL}
    \end{bmatrix}
    \otimes
    \bm{i}
    \otimes
    \left(\prod_{\rm i=k}^{1} {\bf q}_{\rm iL}^{-1}\right)\\
    \otimes
    \left(\prod_{\rm j=1}^{\rm m} {\bf q}_{\rm jR}\right)
    \otimes {\bf q}_{\rm 0R} \quad \text{where } \bm{i}= \text{identity quaternion}
\end{split}
\end{equation}
Writing $\bm{i} = \left(\prod_{\rm i=k+1}^{\rm n} {\bf q}_{\rm iL} \otimes {\bf q}_{\rm 0L}\right) \otimes \left({\bf q}_{\rm 0L}^{-1} \otimes \prod_{\rm i=n}^{\rm k+1} {\bf q}_{\rm iL}^{-1}\right)$ and substituting $\bm{i}$ in~\eqref{eq: partial_kl} and using ${\bf q}_{\rm rel}$ from~\eqref{eq: q_rel_derivation} we get,
\begin{equation}
\label{eq: partial_kl2}
    \frac{\partial {\bf q}_{\rm rel}}{\partial \theta_{\rm kL}} = -\frac{1}{2}
    \left({\bf q}_{\rm 0L}^{-1} \otimes \prod_{\rm i=n}^{\rm k+1} {\bf q}_{\rm iL}^{-1}\right)\otimes
    \begin{bmatrix}
        0 \\ {\bf \omega}_{\rm kL}
    \end{bmatrix}
    \otimes
    \left(\prod_{\rm i=k+1}^{\rm n} {\bf q}_{\rm iL} \otimes {\bf q}_{\rm 0L}\right)
    \otimes
    {\bf q}_{\rm rel}
\end{equation}
Let ${\bf q}_{\rm s} = \left({\bf q}_{\rm 0L}^{-1} \otimes \prod_{\rm i=n}^{\rm k+1} {\bf q}_{\rm iL}^{-1}\right)$, then ${\bf q}_{\rm s}^{-1} = \left(\prod_{\rm i=k+1}^{\rm n} {\bf q}_{\rm iL} \otimes {\bf q}_{\rm 0L}\right)$. Now substituting ${\bf q}_{\rm s}$ in to ~\eqref{eq: partial_kl2} we get,
\begin{eqnarray} \nonumber
\label{eq: partial_kl3}
    &&\frac{\partial {\bf q}_{\rm rel}}{\partial \theta_{\rm kL}} = -\frac{1}{2}{\bf q}_{\rm s}\otimes
    \begin{bmatrix}
        0 \\ {\bf \omega}_{\rm kL}
    \end{bmatrix}
    \otimes
    {\bf q}_s^{-1} \otimes {\bf q}_{\rm rel}\\ \nonumber
    &&= -\frac{1}{2}{\bf q}_s^+{\bf q}_{\rm s}^{-1}
    \begin{bmatrix}
        0 \\ {\bf \omega}_{\rm kL}
    \end{bmatrix}
    \otimes
    {\bf q}_{\rm rel} = -\frac{1}{2} 
    \begin{bmatrix}
        1 & {\bf 0}^{\rm T} \\ {\bf 0} & {\bf R}_{\rm kL}
    \end{bmatrix}
    \begin{bmatrix}
        0 \\  {\bf \omega}_{\rm kL}
    \end{bmatrix}\otimes{\bf q}_{\rm rel}\\
    && = -\frac{1}{2} {\bf q}_{\rm rel}^{\oplus}
    \begin{bmatrix}
    0 \\ {\bf R}_{\rm kL} {\bf \omega}_{\rm kL}
    \end{bmatrix}
    \text{where} \quad {\bf R}_{\rm kL} = {\bf R}_{\rm 0L}^{\rm T}\prod_{\rm i=n}^{\rm k+1} e^{-\hat{\bm{\omega}}_{\rm iL}\theta_{\rm iL}}
    \label{eq: R_kL}
\end{eqnarray}
\noindent
Now we derive a general expression of $\frac{\partial {\bf q}_{\rm rel}}{\partial \theta_{\rm kR}}$ for ${\rm k}=1 \dots {\rm m}$.
\begin{eqnarray} \nonumber
     &&\frac{\partial {\bf q}_{\rm rel}}{\partial \theta_{\rm kR}} = 
    \left({\bf q}_{\rm 0L}^{-1} \otimes \prod_{\rm i=n}^{1} {\bf q}_{\rm iL}^{-1} \otimes \prod_{\rm j=1}^{\rm k-1} {\bf q}_{\rm jR} \right)\otimes \frac{1}{2}
    \begin{bmatrix}
        0 \\ \bm{\omega}_{\rm kR}
    \end{bmatrix}
    \otimes
    \left(\prod_{\rm j=k}^{\rm m} {\bf q}_{\rm jL} \otimes {\bf q}_{\rm 0R}\right)\\
    \label{eq: partial_kR}
    &&= 
    \left({\bf q}_{\rm 0L}^{-1} \otimes \prod_{\rm i=n}^{1} {\bf q}_{\rm iL}^{-1} \otimes \prod_{\rm j=1}^{\rm k-1} {\bf q}_{\rm jR} \right)\otimes \frac{1}{2}
    \begin{bmatrix}
        0 \\ \bm{\omega}_{\rm kR}
    \end{bmatrix}
    \otimes
    \bm{i}
    \otimes
    \left(\prod_{\rm j=k}^{\rm n} {\bf q}_{\rm jL} \otimes {\bf q}_{\rm 0R}\right)\\ \nonumber
    \label{eq: identiy_quaternion}
    &&\text{where,}\\ \nonumber
    &&\bm{i} = \left(\prod_{\rm j=k-1}^{1} {\bf q}_{\rm jR}^{-1} \otimes \prod_{\rm i=1}^{n} {\bf q}_{\rm iL} \otimes {\bf q}_{\rm 0L} \right) \otimes \left({\bf q}_{\rm 0L}^{-1} \otimes \prod_{\rm i=n}^{1} {\bf q}_{\rm iL}^{-1} \otimes \prod_{\rm j=1}^{\rm k-1} {\bf q}_{\rm jR} \right)
\end{eqnarray}
Substituting the value of identity quaternion in~\eqref{eq: identiy_quaternion} in to ~\eqref{eq: partial_kR} we get using the definition of ${\bf q}_{\rm rel}$ and assuming ${\bf q}_{\rm s}$ $=$ $\left(\prod_{\rm j=k-1}^{1} {\bf q}_{\rm jR}^{-1} \otimes \prod_{\rm i=1}^{\rm n} {\bf q}_{\rm iL} \otimes {\bf q}_{\rm 0L} \right)$ as follows,
\begin{eqnarray} \nonumber
    \label{eq: partial_kr2}
    &&\frac{\partial {\bf q}_{\rm rel}}{\partial \theta_{\rm kR}} = \frac{1}{2} {\bf q}_{\rm s} \otimes
    \begin{bmatrix}
        0 \\ \bm{\omega}_{\rm kR}
    \end{bmatrix}
    \otimes
    {\bf q}_{\rm s}^{-1}
    \otimes
    {\bf q}_{\rm rel}\\ \nonumber
    &&=\frac{1}{2} {\bf q}_{\rm s}^+ {\bf q}_{\rm s}^{-1}
    \begin{bmatrix}
        0 \\ {\bf \omega}_{\rm kR}
    \end{bmatrix}
    \otimes
    {\bf q}_{\rm rel}
    =
    \frac{1}{2} {\bf q}_{\rm rel}^{\oplus}
    \begin{bmatrix}
        0 \\ {\bf R}_{\rm kR}{\bf \omega}_{\rm kR}
    \end{bmatrix} \quad \text{where,}\\
    \label{eq: R_kR}
    &&{\bf R}_{\rm kR}=\left({\bf R}_{\rm 0L}^{\rm T}\prod_{\rm i=n}^{1}e^{-\hat{\bm{\omega}}_{\rm iL}\theta_{\rm iL}}\right) \left(\prod_{\rm j=1}^{\rm k-1}e^{\hat{\bm{\omega}}_{\rm jR}\theta_{\rm jR}}\right)
\end{eqnarray}
\eqref{eq: partial_nl2},~\eqref{eq: partial_kl2},~\eqref{eq: partial_kr2}, completely describe columns of $\frac{\partial {\bf q}_{\rm rel}}{\partial \bm{\Theta}}$,
\begin{eqnarray}
\frac{\partial {\bf q}_{\rm rel}}{\partial \bm{\Theta}}= 
\left[
\begin{array}{c|c|c}
\frac{\partial {\bf q}_{\rm rel}}{\partial \theta_{\rm nL}} & \frac{\partial {\bf q}_{\rm rel}}{\partial \theta_{\rm (n-1)L}} \dots \frac{\partial {\bf q}_{\rm rel}}{\partial \theta_{\rm 1L}} & \frac{\partial {\bf q}_{\rm rel}}{\partial \theta_{\rm 1R}} \dots \frac{\partial \bf{q}_{\rm rel}}{\partial \theta_{\rm mR}} \\
 \text{use Eq.}~\eqref{eq: partial_nl2}& \text{use Eq.}~\eqref{eq: partial_kl2} & \text{use Eq.}~\eqref{eq: partial_kr2}
\end{array}
\right]_{4\times(n+m)}
\end{eqnarray}
We can factorize above matrix as $\frac{\partial {\bf q}_{\rm rel}}{\partial \bm{\Theta}} = \frac{1}{2}{\bf H}^{\rm T}{\bf J}_{\rm r}$, where, ${\bf H} = \left[-\bm{\epsilon} \quad \eta {\bf I} + \hat{\bm{\epsilon}}\right]$ such that ${\bf HH}^{\rm T}={\bf I}_{3 \times 3}$ and ${\bf J}_{\rm r} =
    \begin{bmatrix}
        -{\bf \omega}_{\rm nL} & \dots & -{\bf R}_{\rm kL}\bm{\omega}_{\rm kL} & \dots & {\bf R}_{\rm kR}\bm{\omega}_{\rm kR}
    \end{bmatrix} \in \mathbb{R}^{3 \times \rm (n+m)}$
The expressions of ${\bf R}_{\rm kL}$ and ${\bf R}_{\rm kR}$ are as in~\eqref{eq: R_kL} and~\eqref{eq: R_kR} and ${\bf I}$ is $3\times3$ identity matrix. The ${\bf J}_{\rm r}$ matrix here is same as the last three rows of ${\bf J}_{\rm rel}^{\rm s}$ (or ${\bf J}_{\rm rel}^{\rm a}$) as in~\eqref{eq: j_rel_all_col}.
\\

\end{document}